\documentclass{article}
\usepackage{arxiv}

\usepackage[T1]{fontenc}    
\usepackage[utf8]{inputenc}
\usepackage{booktabs}       
\usepackage{amsfonts}       
\usepackage{nicefrac}       
\usepackage{microtype}      

\usepackage{times,soul}
\usepackage{graphicx}
\usepackage{amsmath,amsfonts,bm,mathtools}
\usepackage{adjustbox,booktabs,colortbl}
\usepackage{url,cleveref}
\usepackage{subcaption}
\usepackage{algorithm,algpseudocode}
\usepackage{multirow}

\newcommand{\eg}{\textit{e.g.}}

\newcommand{\ie}{\textit{i.e.}}

\newcommand{\ordinal}[1]{{#1}^\mathrm{th}}
\newcommand{\realset}{\mathbb{R}}
\newcommand{\tx}{\mathbf{x}}
\newcommand{\tz}{\mathbf{z}}
\newcommand{\sign}{\mathrm{sign}}
\newcommand{\clip}{\mathrm{clip}}
\newcommand{\pool}{\mathrm{pool}}

\graphicspath{{images/}}

\title{Sitatapatra: Blocking the Transfer of Adversarial Samples}
\author{%
Ilia Shumailov\thanks{
    Ilia Shumailov, Xitong Gao and Yiren Zhao
    contributed equally to this work.
}\\ University of Cambridge \And%
Xitong Gao\footnotemark[1]\hspace{4pt}\\ Shenzhen Institutes of Advanced Technology\And%
Yiren Zhao\footnotemark[1]\hspace{4pt}\\ University of Cambridge \And%
Robert Mullins\\University of Cambridge \And%
Ross Anderson\\University of Cambridge \And%
Cheng-Zhong Xu\\University of Macau \And%
}
\begin{document}








\maketitle

\begin{abstract}
Convolutional Neural Networks (CNNs)
are widely used to solve classification tasks in computer vision.
However, they can be tricked into misclassifying specially
crafted ``adversarial'' samples -- and samples built to trick one model
often work alarmingly well against other models trained on
the same task. In this paper we introduce Sitatapatra,
a system designed to block the transfer of adversarial samples.
We show how Sitatapatra has the potential to serve as a key and
diversifies neural networks, as in
cryptography, and provides a mechanism for detecting attacks.
What's more, when adversarial
samples are detected they can typically be traced back to the individual
device that was used to develop them.
The run-time overheads are minimal permitting the use of Sitatapatra on
constrained systems.

\end{abstract}

\section{Introduction}
Convolutional Neural Networks (CNNs)
have achieved revolutionary
performance in vision-related
tasks such as
segmentation \cite{badrinarayanan2015segnet},
image classification \cite{krizhevsky2012imagenet},
object detection \cite{ren2015faster}
and
visual question answering \cite{antol2015vqa}.
It is now inevitable that CNNs will be deployed widely for a broad range of applications including both safety-critical and security-critical tasks such as
autonomous vehicles \cite{fang2003road},
face recognition \cite{schroff2015facenet}
and action recognition \cite{ji20133d}.

However, researchers have discovered that small perturbations
to input images can trick CNNs ---
with classifiers producing results surprisingly far from the correct answer
in response to small perturbations
that are not perceptible by humans.
An attacker can thus create adversarial image inputs
that cause a CNN to misbehave.
The resulting attack scenarios are broad,
ranging from breaking into smartphones
through face-recognition systems \cite{Carlini2016}
to misdirecting autonomous vehicles
through perturbed road signs \cite{eykholt2018robust}.

The attacks are also practical, because they are surprisingly portable.
If devices are shipped with the same CNN classifier,
the attacker only needs to analyze a single one of them to
perform effective attacks on the others.
Imagine a firm that deploys the same CNN on devices in
very different environments.
An attacker might not be able to experiment with a security
camera in a bank vault, but if a cheaper range of
security cameras are sold for home use and use the same CNN, he may buy
one and generate transferable adversarial samples using it. What is more, making
CNNs sufficiently different is not straightforward. Recent
research teaches that different CNNs trained to solve similar problems
are often vulnerable to the same adversarial samples~\cite{papernot2016transferability,zhao2018compress}. It is clear that we need an efficient way of
limiting the transferability of adversarial samples.
And given the complexity of the scenarios in which sensors operate,
the protection mechanism should:
\emph{a)} work in computationally constrained environments;
\emph{b)} require minimal changes to the model architecture, and;
\emph{c)} offer a way of building complex security policies.

In this paper we propose Sitatapatra, a system designed and built
to constrain the transferability of adversarial examples.
Inspired by cryptography, we introduce a notion of \textit{key} into CNNs that
causes each network of the same architecture to be internally
different enough to stop the transferability of adversarial attacks.
We describe multiple ways of embedding the \textit{keys}
and evaluate them extensively on a range of computer vision benchmarks.
Based on these data, we propose a scheme to pick \textit{keys}.
Finally, we discuss the scalability of the Sitatapatra defence and
describe the trade-offs inherent in its design.

The contributions of this paper are:
\begin{itemize}
    \item We introduce Sitatapatra, the first system designed to stop
    adversarial samples being transferable.
    \item We describe how to embed a secret \textit{key}
    into CNNs.
    \item We show that Sitatapatra not only blocks sample transfer,
    but also allows detection and attack attribution.
    \item We measure performance and show that the run-time computational overhead is low enough (0.6--7\%) for Sitatapatra to work on constrained devices.
\end{itemize}

The paper is structured as follows.
\Cref{sec:rw} describes the related work.
\Cref{sec:meth} introduces
the methodology of Sitatapatra and presents the design choices.
\Cref{sec:eval} evaluates the system against state-of-the-art attacks on three popular image classification benchmarks.
\Cref{sec:disc} discusses the trade-offs in Sitatapatra,
analyses \textit{key} diversity and the costs of \textit{key} change, and
discusses
how to deploy Sitatapatra effectively at scale.

\section{Related Work}
\label{sec:rw}
Since the invention of adversarial attacks
against CNNs \cite{Szegedy2013},
there has been rapid co-evolution of attack and defense techniques
in the deep learning community.

An adversarial sample is defined as
a slightly perturbed image \( \hat\tx \)
of the original \( \tx \),
while \( \hat\tx \) and \( \tx \)
are assigned different classes
by a CNN classifier \( F \).
The \( l^p \)-norm of the perturbation
is often constrained by a small constant \( \epsilon \)
such that \( { \| \hat\tx - \tx \| }_p \leq \epsilon \)
and \( 0 \leq \hat\tx \leq 1 \),
where \( p \) can be 1, 2, or \( \infty \).

The fast gradient sign method (FGSM)~\cite{goodfellow2014explaining}
is a simple and effective \( L^\infty \) attack
for finding such samples.
FGSM generates \( \hat\tx \)
by computing the gradient of the target class \( y \)
with respect to \( \tx \),
and applies a distortion \( \epsilon \)
for all pixels against the sign of the gradient direction:
\begin{equation}
    \hat\tx = \clip \left(
        \tx + \epsilon \cdot \sign \left(
            \nabla_\tx \ell_y \left(F \left( x \right) \right)
        \right)
    \right),
\end{equation}
where \( \ell_y(F(\tx)) \) denotes
the loss of the network output \( F(\tx) \)
for the target class \( y \),
\( \nabla_\tx (\tz) \) evaluates
the gradient of \( \tz \) with respect to \( \tx \),
the \( \sign \) function returns
the signs \( \{ \pm1 \} \) of the values in its input tensor,
and finally \( \clip( \tz ) = \max(\min(\tz, 1), 0) \)
constraints each value to the range of permissible pixel values,
\ie~\( [0, 1] \). An iterative version of the attack, Projected Gradient Descent (PGD), was later proposed~\cite{kurakin2016adversarial}.

DeepFool~\cite{Moosavi15} is an attack
that iteratively linearizes
misclassification boundaries of the network,
and perturbs the image by moving along the direction
that gives the nearest misclassification.

The Carlini \& Wagner attack~\cite{carlini2017towards} formulates
the following optimization problem,
whose solution gives an adversarial sample:
\begin{equation}
    \min_{\tx^\star}
        { \left\| \tx^\star - \tx \right\| }^2_2 +
        c \cdot G(\tx^\star).
\end{equation}
Here, the first term optimizes the \( L^2 \)-distance,
while the second term
\(
    G(\tx^\star) = \max (
        \{ -\kappa \} \cup
        \{
            \ell_{y^\prime}( \tz ) - \ell_y( \tz )
        \mid
            y^\prime \neq y, \tz = F(\tx^\star)
        \}
    )
\)
minimizes the loss of classes other than \( y \),
and \( \kappa \) controls the confidence of misclassification.

Furthermore, it is well-known that
adversarial samples demonstrate good transferability~\cite{
    Szegedy2013, goodfellow2014explaining, liu2016delving},
\ie~such samples generated from a model \( F \)
tend to remain adversarial for models other than \( F \).
This is problematic as black-box attacks
may generate strong adversarial inputs,
without even requiring any prior knowledge
of the architecture and training procedures
of the model under attack~\cite{PapernotMGJCS16}.
\Cref{tab:pretrain} shows that exact scenario ---
adversarial samples generated from one model
can effectively transfer
to another with the same architecture
trained only with different initializations.

Many researchers have proposed defences based on changes
to the network architecture or the training procedure.
Random self-ensemble~\cite{liu2018rse}
adds noise to computed features
to simulate a random ensemble of models.
Deep Defence~\cite{yan2018deepdef}
incorporates the perturbation
required to generate adversarial samples
as a regularization during training,
thus making the trained model
less susceptible to adversarial samples.
Other methods such as
adversarial training~\cite{goodfellow2014explaining},
defensive distillation~\cite{papernot2016distill} and
Bayesian neural networks~\cite{liu2018advbnn},
also demonstrate good resistance against
adversarial samples.
The above methods can produce robust models
with greater attack resistance,
yet this is often at the cost of
the original model's accuracy~\cite{tsipras2018robustness}.
For this reason, many other researchers extended the networks
to detect adversarial attacks explicitly~\cite{metzen2017detecting, meng2017magnet, shumailov2018taboo}, but some of these detection methods can add considerable
computational overhead.

\begin{table*}[]
\centering

\begin{adjustbox}{width=\textwidth,center}
\begin{tabular}{@{}ll|lllll|lllll|lllll@{}}
\toprule
&
& \multicolumn{5}{c}{LeNet5}
& \multicolumn{5}{c}{MCifarNet}
& \multicolumn{5}{c}{ResNet18-Cifar10}
\\
& Params
&$S_{\mathsf{Acc}}(\%)$     &$T_{\mathsf{Acc}}(\%)$    &$\Delta_{\mathsf{Acc}}(\%)$   &$l_2$   &$l_{\infty}$
&$S_{\mathsf{Acc}}(\%)$     &$T_{\mathsf{Acc}}(\%)$    &$\Delta_{\mathsf{Acc}}(\%)$   &$l_2$   &$l_{\infty}$
&$S_{\mathsf{Acc}}(\%)$     &$T_{\mathsf{Acc}}(\%)$    &$\Delta_{\mathsf{Acc}}(\%)$   &$l_2$   &$l_{\infty}$
\\ \midrule
Clean&
&99.13    &99.25    &-             &-       &-
&89.51    &90.14    &-             &-       &-
&91.12    &93.58    &-             &-       &-
\\ \midrule
\multirow{3}{*}{FGSM}
& $\epsilon=0.02$
&98.20 &98.91 &0.70 &0.42 &0.02
&17.03 &24.22 &7.19 &0.98 &0.02
&18.05 &50.16 &32.11 &0.99 &0.02
\\
& $\epsilon=0.06$
&92.19 &96.17 &3.98 &1.65 &0.08
&2.89 &11.17 &8.28 &3.67 &0.07
&4.61 &22.66 &18.05 &3.71 &0.07
\\
& $\epsilon=0.6$
&0.55 &3.59 &3.05 &11.91 &0.59
&1.02 &7.81 &6.80 &20.98 &0.48
&0.00 &8.59 &8.59 &22.07 &0.50
\\
\midrule
\multirow{2}{*}{DeepFool}
& $i=50$
&14.92 &92.03 &77.11 &1.86 &0.44
&1.02 &27.34 &26.33 &0.20 &0.03
&1.02 &88.20 &87.19 &0.16 &0.02
\\
& $i=10$
&87.73 &96.95 &9.22 &1.76 &0.41
&1.80 &27.34 &25.55 &0.20 &0.03
&1.72 &88.20 &86.48 &0.16 &0.02
\\
\midrule
\multirow{2}{*}{C\&W}
&$lr=0.5$
&3.75 &86.56 &82.81 &3.09 &0.71
&2.03 &10.47 &8.44 &9.86 &0.24
&0.55 &15.00 &14.45 &10.02 &0.24
\\
&$lr=1.0$
&1.02 &60.86 &59.84 &4.25 &0.76
&5.00 &11.02 &6.02 &17.28 &0.44
&0.62 &12.81 &12.19 &17.75 &0.45
\\ \midrule
C\&W
&$lr=0.5$
&3.91 &15.78 &11.88 &5.12 &0.90
&2.73 &4.69 &1.95 &15.89 &0.52
&1.09 &10.39 &9.30 &14.63 &0.43
\\
$c=0.999$
&$lr=1.0$
&1.09 &6.17 &5.08 &5.75 &0.95
&5.47 &5.23 &-0.23 &24.27 &0.77
&1.48 &8.83 &7.34 &25.35 &0.81
\\
\bottomrule
\end{tabular}
\end{adjustbox}
\caption{
    Transferability of two pretrained models with different
    initializations on various datasets.
    Source and Target are two models of the same topology
    but different initialization point.
    $S_{\mathsf{Acc}}$ and $T_{\mathsf{Acc}}$ are the accuracies on
    source and target models.
    $l_2$ and $l_{\infty}$ are the norm values of the adversarial noise,
    larger norm values mean larger perturbations.
    $\epsilon$ is a hyperparameter in FGSM,
    $i$ is the number of iterations and $lr$ is the learning rate.
    $c$ is the confidence, the adversarial sample has to cause
    the misclassified class to have a probability above this given threshold.
    }
\label{tab:pretrain}
\end{table*}

\section{Method}\label{sec:meth}

\subsection{High-Level View}\label{sec:meth:hlv}

\begin{figure}[!ht]
    \centering
    \begin{subfigure}{.5\textwidth}
        \includegraphics[width=.9\linewidth]{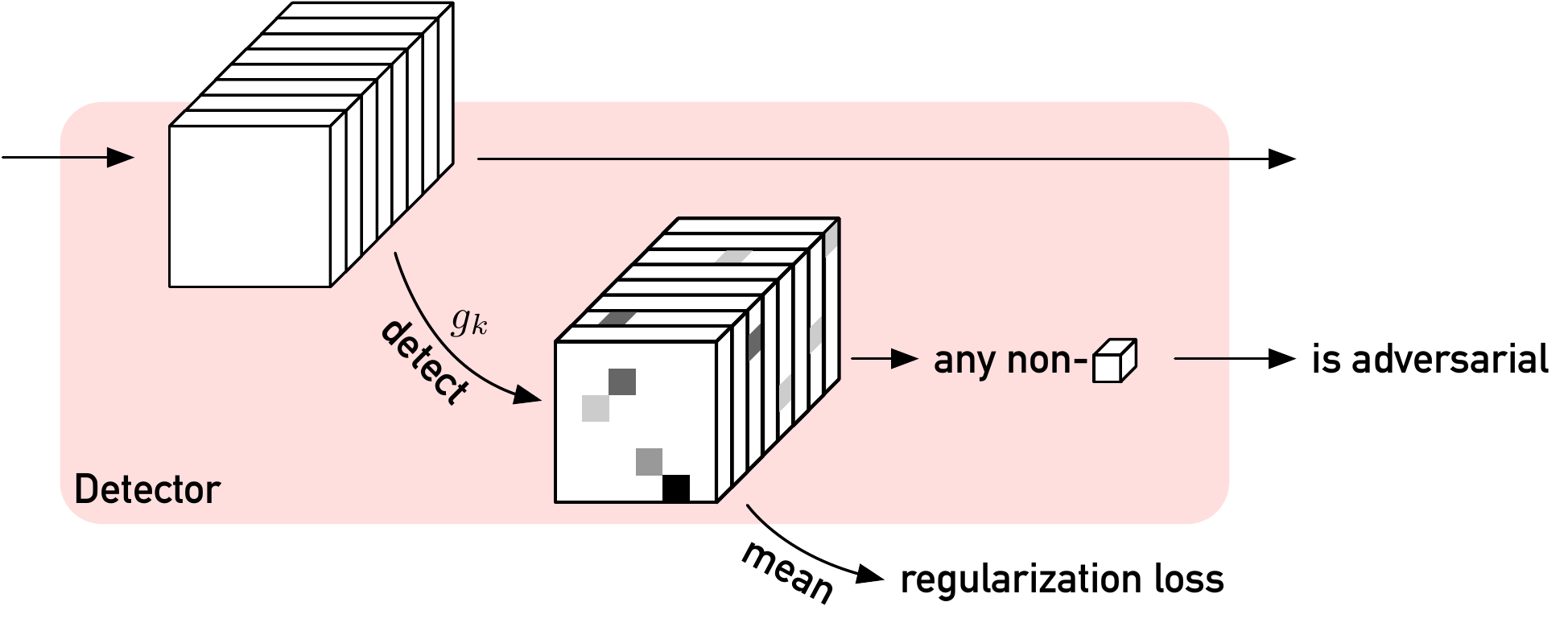}
        \caption{Detector.}\label{fig:detector}
    \end{subfigure}\hfill
    \begin{subfigure}{.5\textwidth}
        \includegraphics[width=.9\linewidth]{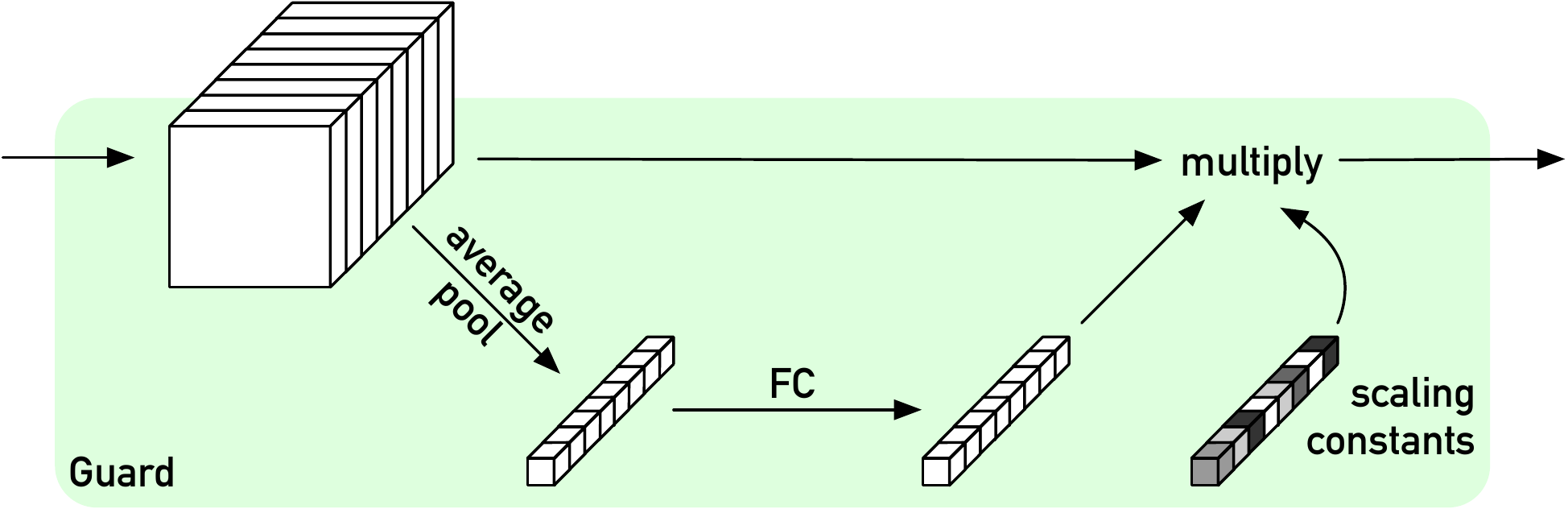}
        \caption{Guard.}\label{fig:guard}
    \end{subfigure}
    \caption{%
        High-level view of the module extensions
        we add to networks
        to stop and detect transferred adversarial samples.
    }
\end{figure}

We start with a high-level description of our method.
Each convolutional layer with ReLU activation
is sequentially extended with a guard (\Cref{fig:guard})
and a detector (\Cref{fig:detector}) layer.
Intuitively, the guard encourages the gradient to disperse
among differently initialized models,
limiting sample transferability.
If this fails, the detector works as our second line of defence
by raising an alarm at potentially adversarial samples.
The rest of this section explains the design of the two modules,
which work in tandem or individually to
defend against and/or detect most instances of the adversarial attacks
considered here.

\subsection{Static Channel Manipulation}%
\label{sec:meth:det}

Where an attack sample is transferable, we hypothesize that
by moving along the gradient direction
of the input image that gave rise to the adversarial sample,
some of the neurons of the network are likely to be highly excited.
By monitoring abnormal excitation,
we can detect many adversarial samples.

In Sitatapatra, we use 
a simple concept -- we train neural networks
using the original datasets,
but with additional activation regularization, so that a certain
set of outputs, and of intermediate activation values, are not
generated by any of the training set inputs. If one of these `taboo'
activations is observed, this serves as a signal that the input may be
adversarial. As different instances of the model can be trained
with different taboo sets, we have a way to introduce diversity that is
analogous to the \textit{keys} used in cryptographic systems.
This approach is much faster
than, for example, adversarial training~\cite{goodfellow2014explaining}.
By regularizing the activation feature maps,
clean samples yield well-bounded activation results,
yet adversarial ones typically result in
highly excited neuron outputs in activations~\cite{shumailov2018taboo}.
Using this,
we can design our detector (\Cref{fig:detector})
so that uncompromised models
can successfully block most adversarial samples
transferred from compromised models,
even under the strong assumption
that the models share
the same network architecture and dataset.

For simplicity, we consider a feed-forward CNN \( F \)
consisting of a sequence of \( L \) convolutional layers,
where the \( \ordinal{l} \) layer
computes output feature maps \(
    \tx_{l} \in \realset^{C_l \times H_l \times W_l}
\).
Here, \( \tx_{l} \) is a collection of feature maps
with \( C_l \) channels of \( H_l \times W_l \) images,
where \( H_l \) and \( W_l \)
respectively denote the width and height of the image.

The detector module additionally performs
the following polynomial transformation \( f_k \)
to the features of each layer output,
which can be derived
from the \textit{key} of the network:
\begin{equation}
    f_k(\tx_l) =
        a_n \tx_l^{n} + a_{n-1} \tx^{n-1} + \cdots + a_0,
\end{equation}
and the transformation is then followed
by the following criterion evaluation:
\begin{equation}
    g_k(\tx_l) =
        f_k\left(\tx_l\right) \cdot
        {\left(\sign\left(
            f_k\left(\tx_l\right) - t
        \right)\right)}_+.
\end{equation}
Here,
\( \cdot \) is element-wise multiplication,
\( {(\tz)}_+ = \max(0, \tz) \)
denotes the ReLU activation,
the degrees of the polynomial \( n \)
and the constant values
\( a_0 \), \( a_1 \), \ldots, \( a_n \) and \( t > 0 \)
can be identified
as the \textit{key} embedded into the network.
During training with clean dataset samples,
any values \( f_k(\tx_l) > t \) are penalized,
with the following regularization term:
\begin{equation}
    \mathcal{R}(\tx) =
        \sum_{l \in L} {\left\|
            g_k\left(\tx_l\right)
        \right\|}_1.
\end{equation}

As we will examine closely in \Cref{sec:eval},
the \( g_k(\tx_l) \) evaluated
from adversarial samples
generally produce positive results,
whereas clean validation samples
usually do not trigger this criterion.
Our detector uses this to identify adversarial samples.
In addition, the polynomial can be evaluated efficiently
with Horner's method~\cite{knuth1962evaluation},
requiring only
\(n\) multiply-accumulate operations per value,
which is insignificant when compared to
the computational cost of the convolutions.
Finally,
the non-linear nature of our regularization
diversifies the weight distributions among models,
which intuitively explains why adversarial samples
become less likely to transfer.

\subsection{Dynamic Channel Manipulation}%
\label{sec:meth:stop}

In this section, we present the design of the guard module.
It dynamically manipulates feature maps using per-channel attention,
which is in turn inspired by
the squeeze-excitation networks~\cite{hu2017squeeze} and feature boosting~\cite{gao2018dynamic}.
As with the detector modules
described earlier in \Cref{sec:meth:det},
we extend existing networks
by adding a guard module immediately after
each convolutional layer and its detector module.
The \( \ordinal{l} \) module accepts \( \tx_l \)
computed by the previous convolutional layer as its input,
and introduces a small auxiliary network
\(
    h: \realset^{C_l \times H_l \times W_l} \to
       \realset^{C_l \times H_l \times W_l}
\),
which amplifies important channels
and suppresses unimportant ones in feature maps,
before feeding the results of \( h(\tx_l) \)
to the next convolutional layer.
The auxiliary network
is illustrated in \Cref{fig:guard}
and is defined as:
\begin{equation}
    h(\tx_l) = \tx_l \cdot
        \left(
            \bm\gamma_l \cdot \pool\left( \tx_l \right) \bm\phi_l
        \right),
\end{equation}
where \(
    \pool: \realset^{C_l \times H_l \times W_l} \to
           \realset^{C_l}
\) performs a global average pool
on the feature maps \( \tx_l \)
which reduces them to a vector of \( C_l \) values,
\( \bm\phi \in \realset^{C_l \times C_l} \)
is a matrix of trainable parameters,
and \( \cdot \)
denotes element-wise multiplication between tensors
which broadcasts in a channel-wise manner.
Finally, \( \bm\gamma_l \) is a vector of
\( C_l \) scaling constants,
where each value is randomly drawn
from a uniform distribution
between 0 and 1,
using the crypto-key embedded within the model
as the random seed.
By doing so,
each deployed model can have
a different \( \bm\gamma_l \) value,
which diversifies the gradients among models
and forces the fine-tuned models
to adopt different local minima.


\section{Evaluation}\label{sec:eval}

\subsection{Networks, Datasets and Attacks}
We evaluate Sitatapatra on two datasets:
MNIST \cite{lecun2010mnist},
CIFAR10 \cite{krizhevsky2014cifar}.
In MNIST,
we use the LeNet5 architecture \cite{lecun2015lenet}.
In CIFAR10,
we use an efficient CNN architecture
(MCifarNet) from Mayo \cite{zhao2018mayo}
that achieved a high classification rate using only 1.3M parameters.
ResNet18
is also considered on the CIFAR10 dataset~\cite{he2016deep}.

We evaluated the performance of clean
networks and Sitatapatra networks, using two clean models with different
initializations as the baseline.
For the Sitatapatra models we had two
different keys applied on the source and target models. The keys used are
$2x^2+3x+5<6$ and $0.1x^2-x+2<3$ respectively. 
Throughout the development of Sitatapatra we have tried using a large number of different keys and observed that the performance of the detector is largely dependant on the strictness of the regulariser. 
For evaluation purposes we have decided to report on the worst performance, practically showing the detection lower bound. 
The relative performance of other keys will be presented later in \Cref{sec:disc}.
We consider attacks listed in~\Cref{sec:rw} in White and Black-box settings. For the latter, we estimate gradients using a simple coordinate-wise finite difference method~\cite{DBLP:journals/corr/abs-1712-09196}.

\subsection{Static and Dynamic Channel Manipulations}

\begin{table*}[]
\centering

\begin{adjustbox}{width=\textwidth,center}
\begin{tabular}{@{}ll|llllll|llllll|llllll@{}}
\toprule
&
& \multicolumn{6}{c}{LeNet5}
& \multicolumn{6}{c}{MCifarNet}
& \multicolumn{6}{c}{ResNet18-Cifar10} \\
& Params
&$S_{\mathsf{Acc}}(\%)$     &$T_{\mathsf{Acc}}(\%)$    &$\Delta_{\mathsf{Acc}}(\%)$  &$T_{\mathsf{Det}}(\%)$ &$l_2$   &$l_{\infty}$
&$S_{\mathsf{Acc}}(\%)$     &$T_{\mathsf{Acc}}(\%)$    &$\Delta_{\mathsf{Acc}}(\%)$  &$T_{\mathsf{Det}}(\%)$ &$l_2$   &$l_{\infty}$
&$S_{\mathsf{Acc}}(\%)$     &$T_{\mathsf{Acc}}(\%)$    &$\Delta_{\mathsf{Acc}}(\%)$  &$T_{\mathsf{Det}}(\%)$ &$l_2$   &$l_{\infty}$
\\ \midrule
Clean&
&99.13    &99.25    &-             &-       &-            &-
&89.51    &90.14    &-             &-       &-            &-
&91.12    &93.58    &-             &-       &-            &-
\\ \midrule
\multirow{5}{*}{FGSM}
& $\epsilon=0.02$
&97.30 &98.44 &1.14 &0.00 &0.43 &0.02
&26.42 &38.64 &12.22 &1.85 &1.00 &0.02
&39.29 &59.41 &20.12 &1.15 &1.03 &0.02
\\
& $\epsilon=0.05$
&92.19 &97.44 &5.26 &0.00 &1.07 &0.05
&22.87 &32.95 &10.09 &1.53 &2.43 &0.04
&30.18 &44.74 &14.56 &0.75 &2.50 &0.05
\\
& $\epsilon=0.1$
&80.11 &95.45 &15.34 &0.00 &2.13 &0.10
&16.48 &25.00 &8.52 &3.17 &4.63 &0.09
&19.28 &30.41 &11.13 &20.10 &4.84 &0.09
\\
& $\epsilon=0.15$
&63.07 &90.06 &26.99 &0.00 &3.18 &0.15
&9.23 &17.47 &8.24 &18.11 &6.84 &0.13
&11.51 &24.01 &12.50 &93.02 &7.12 &0.13
\\
& $\epsilon=0.35$
&6.68 &30.54 &23.86 &0.00 &7.31 &0.35
&1.14 &14.20 &13.07 &86.29 &14.61 &0.29
&3.51 &13.57 &10.06 &91.69 &15.21 &0.30
\\
\midrule
\multirow{5}{*}{FGSM /w GE}
& $\epsilon=0.02$
&97.87 &99.15 &1.28 &0.00 &0.43 &0.02
&28.41 &39.35 &10.94 &2.03 &0.97 &0.02
&40.91 &60.94 &20.03 &1.68 &1.03 &0.02
\\
& $\epsilon=0.05$
&95.31 &98.44 &3.12 &0.00 &1.07 &0.05
&25.28 &35.65 &10.37 &2.09 &2.38 &0.04
&29.69 &44.32 &14.63 &0.25 &2.50 &0.05
\\
& $\epsilon=0.1$
&82.39 &96.31 &13.92 &0.00 &2.14 &0.10
&18.04 &26.56 &8.52 &3.28 &4.47 &0.08
&18.32 &29.40 &11.08 &17.63 &4.88 &0.09
\\
& $\epsilon=0.15$
&62.93 &92.61 &29.69 &0.00 &3.19 &0.15
&11.22 &19.32 &8.10 &17.24 &6.54 &0.12
&9.94 &23.44 &13.49 &93.85 &7.17 &0.13
\\
& $\epsilon=0.35$
&7.67 &32.10 &24.43 &0.00 &7.35 &0.35
&1.42 &15.48 &14.06 &84.84 &14.21 &0.28
&2.70 &14.63 &11.93 &91.86 &15.27 &0.30
\\
\midrule
\multirow{3}{*}{PGD}
& $i=10$
&64.63 &98.44 &33.81 &0.00 &1.57 &0.09
&1.14 &41.90 &40.77 &1.84 &0.76 &0.02
&12.12 &61.09 &48.97 &0.30 &0.82 &0.03
\\
& $i=20$
&0.85 &98.44 &97.59 &0.00 &1.96 &0.12
&0.28 &41.90 &41.62 &1.84 &0.76 &0.02
&10.40 &61.09 &50.69 &0.30 &0.90 &0.04
\\
& $i=50$
&0.00 &98.44 &98.44 &0.00 &1.96 &0.12
&0.00 &41.90 &41.90 &1.84 &0.76 &0.02
&7.43 &60.94 &53.51 &0.29 &1.03 &0.05
\\
\midrule
\multirow{3}{*}{PGD /w GE}
& $i=10$
&65.06 &98.72 &33.66 &0.00 &1.61 &0.09
&0.99 &43.61 &42.61 &1.89 &0.76 &0.02
&13.21 &63.21 &50.00 &1.04 &0.83 &0.03
\\
& $i=20$
&0.85 &98.72 &97.87 &0.00 &1.99 &0.12
&0.28 &43.61 &43.32 &1.89 &0.76 &0.02
&11.22 &63.21 &51.99 &1.04 &0.92 &0.04
\\
& $i=50$
&0.00 &98.72 &98.72 &0.00 &2.00 &0.12
&0.00 &43.61 &43.61 &1.89 &0.76 &0.02
&8.24 &63.07 &54.83 &1.04 &1.06 &0.05
\\
\midrule
\multirow{2}{*}{DeepFool}
& $i=50$
&11.48 &98.91 &87.42 &0.00 &1.51 &0.39
&0.23 &76.25 &76.02 &1.30 &0.38 &0.05
&3.75 &77.27 &73.52 &7.63 &0.87 &0.10
\\
& $i=10$
&89.61 &99.22 &9.61 &0.00 &1.44 &0.37
&2.97 &76.17 &73.20 &1.30 &0.38 &0.05
&6.33 &77.50 &71.17 &6.95 &0.81 &0.10
\\
\midrule
\multirow{2}{*}{C\&W}
&$lr=0.5$
&4.38 &93.98 &89.61 &2.35 &2.95 &0.66
&3.83 &11.48 &7.66 &67.41 &9.82 &0.25
&2.73 &20.55 &17.81 &95.50 &10.11 &0.25
\\
&$lr=1.0$
&1.56 &80.47 &78.91 &4.78 &3.74 &0.75
&5.31 &13.91 &8.59 &92.02 &17.11 &0.43
&3.52 &11.72 &8.20 &96.05 &17.84 &0.46
\\
\midrule
C\&W
&$lr=0.5$
&4.61 &29.92 &25.31 &16.02 &5.22 &0.94
&4.53 &4.77 &0.23 &99.10 &17.37 &0.61
&3.05 &15.70 &12.66 &98.60 &12.63 &0.38
\\
$c=0.999$
&$lr=1.0$
&1.88 &14.53 &12.66 &24.07 &5.92 &0.98
&7.19 &7.58 &0.39   &100.00 &23.58 &0.80
&5.55 &8.67 &3.12 &98.72 &22.20 &0.72
\\
\bottomrule
\end{tabular}
\end{adjustbox}
\caption{
    Transferability of different models on various datasets
    with the SDCM instrumentation (Guard and Detector).
    Source and Target are two models of the same topology but different
    SDCM instrumentations.
    We generate adversarial samples from source models and apply them
    on the target models.
    $S_{\mathsf{Acc}}$ and $T_{\mathsf{Acc}}$ are the accuracies on
    source and target models.
    $T_{\mathsf{Det}}$ is the detection ratio on the targeting model.
    $l_2$ and $l_{\infty}$ are the norm values of the adversarial noise,
    larger norm values mean larger perturbations.
    $\epsilon$ is a hyperparameter in FGSM,
    $i$ is the number of iterations and $lr$ is the learning rate.
    $c$ is the confidence, the adversarial sample has to cause
    the misclassified class to have a probability above this given threshold. `w/ GE' indicates attacks that use numerical gradient estimation, and they are only reported on MCifarNet because of their long run-time nature.
    }
\label{tab:combined}
\end{table*}

\Cref{tab:combined} shows the performance of combined
static and dynamic channels manipulation (SDCM) method.
During the training phase we tune each model to have
less than $1\%$ false positive detection rate on the clean
evaluation dataset.
The results show good performance on large models, since
large models have more channels available for the proposed instrumentations.
For ResNet18 on CIFAR10,
we can achieve above $95\%$ detection of all adversarial samples.
Similarly, we believe that the performance increase,
in comparison to both LeNet5 and MCifarNet,
comes from an increased number of channels of the ResNet18 model.
Furthermore,
we observe a trade-off between the accuracy difference ($\Delta_{\mathsf{acc}}$)
and the detection rate ($T_{\mathsf{det}}$).
When attacks achieve low classification accuracies on the target model,
the detector, acting as a second line of defence, usually identifies the
adversarial examples.

Further, it can be noticed that the samples that actually transfer well usually
result in relatively larger distortions --- \eg~C\&W with high confidence and FGSM \( 0.6 \).
Although large distortions allow attacks to trick the models,
they inevitably trigger the alarm and thus get detected.
Meanwhile, attacks with more fine-grained perturbations
fail to transfer and the accuracy
$T_{\mathsf{acc}}$ remains high.
For example, DeepFool consistently shows high $T_{\mathsf{acc}}$ and high
$\Delta_{\mathsf{acc}}$.

\begin{table}[!h]
\centering
\begin{adjustbox}{scale=0.8,center}
\begin{tabular}{@{}ll|l|ll|ll@{}}
\toprule
& \multicolumn{1}{c}{}
& \multicolumn{1}{c}{Pretrained}
& \multicolumn{2}{c}{Static Only}
& \multicolumn{2}{c}{Combined}       \\
&
&$\Delta_{\mathsf{Acc}}(\%)$
&$\Delta_{\mathsf{Acc}}(\%)$    &$T_{\mathsf{Det}}(\%)$
&$\Delta_{\mathsf{Acc}}(\%)$    &$T_{\mathsf{Det}}(\%)$
\\ \midrule
\multirow{3}{*}{FGSM}
 & $\epsilon=0.02$
 &0.70 &1.87  &0.00 &0.94 &0.00 \\
 & $\epsilon=0.08$
 &3.98 &15.78 &0.00 &9.14 &0.00 \\
 & $\epsilon=0.6$
 &3.05 &10.78 &0.19 &11.02 &22.28 \\
 \midrule
\multirow{2}{*}{DeepFool}
 & $i=50$
 &77.11 &78.20 &0.00 &87.42 &0.00 \\
 & $i=10$
 &9.22  &1.02  &0.00 &9.61 &0.00 \\
 \midrule
C\&W
 &$lr=0.5$
 &82.81 &91.02 &0.00 &89.61 &2.35 \\
 &$lr=1.0$
 &59.84 &90.39 &0.95 &78.91 &4.78 \\
 \midrule
$c=0.999$
 &$lr=0.5$
 &11.88 &57.50 &2.38 &25.31 &16.02 \\
 &$lr=1.0$
 &5.08  &45.00 &6.47 &12.66 &24.07 \\
\bottomrule
\end{tabular}
\end{adjustbox}
\caption{
    Transferability of LeNet5 on MNIST with different instrumentations.
    $\Delta_{\mathsf{acc}}$ is the difference
    in accuracies and $T_{\mathsf{det}}$ is the detection rate.
    $l_2$ and $l_{\infty}$ are
    the averaged $l_2$ and $l_{\infty}$ norms of
    the distortions on all input images.
    $\epsilon$ is a hyperparameter in FGSM,
    $i$ is the number of steps and
    $lr$ is learning rate.
    $c$ is the confidence, the adversarial sample has to cause
    the misclassified class to have a probability above this given threshold.
    We use $20$ steps on C\&W.}
\label{tab:compare}
\end{table}


Finally, to demonstrate the overall performance of Sitatapatra, we
present a comparison between the baseline, SCM and SDCM in \Cref{tab:compare}.
Both of the proposed methods perform relatively worse for LeNet5
, as mentioned before,
we believe that this due to the small number of channels at its disposal.
Having said that, the proposed methods
still greatly outperforms the baseline in terms of $\Delta_{\mathsf{acc}}$.
The results of DeepFool are very similar across all three methods.
DeepFool was not designed with transferability in mind.
In contrast, C\&W was previously shown to generate
highly transferable adversarial samples.
The combined method manages to successfully
classify the transferred samples and shows sensible detection rates.

\section{Discussion}\label{sec:disc}

\subsection{Accuracy, Detection and Computation Overheads}

In this section we discuss the trade-offs facing the Sitatapatra user.
First and foremost,
the choice of keys has an impact on transferability.

\begin{figure}[!h]
\centering
\begin{subfigure}{0.4\linewidth}
  \centering
  \includegraphics[width=\linewidth]{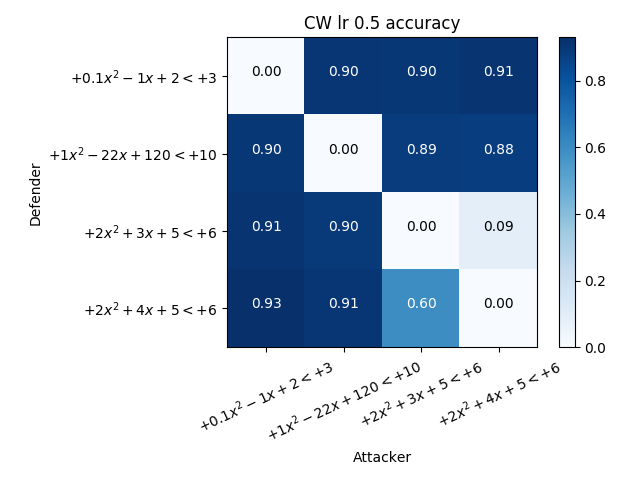}
  \caption{Change in Accuracies ($\Delta_{\mathsf{acc}}$).}
  \label{fig:cwlr05_acc}
\end{subfigure}
\hfill
\begin{subfigure}{0.4\linewidth}
  \centering
  \includegraphics[width=\linewidth]{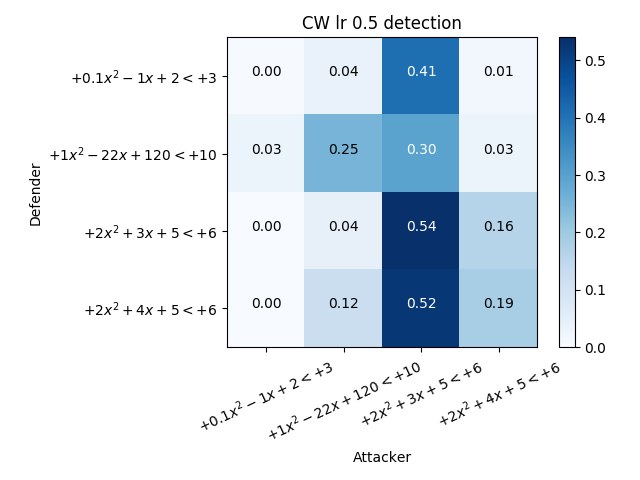}
  \caption{Detection rates on the target model ($T_{\mathsf{det}}$)}
  \label{fig:cwlr05_det}
\end{subfigure}
\caption{Changes in accuracies and detection rates of LeNet5 with various SDCM instrumentations}
\label{fig:confusion}
\end{figure}

In \Cref{fig:confusion}, we present a confusion matrix
with differently instrumented LeNet5 networks
attacking each other with C\&W attack for $20$ steps and a learning rate of $0.5$.
The chosen instrumentations are all second-order polynomials,
but with different coefficients.
It is apparent that Sitatapatra improves the accuracies
reported in \Cref{fig:cwlr05_acc}.
However, for the two polynomials, $2x^2+3x+5<6$ and $2x^2+4x+5<6$,
the behavior is different.
When using the adversarial samples from $2x^2+4x+5<6$ to attack $2x^2+3x+5<6$,
there is almost zero change in
accuracy between the source and the target model, but this relationship
does not hold in reverse.
The above phenomenon indicates that
the adversarial samples generated from one polynomial
are transferable to the other when the polynomials are too similar.
The same effect is observed in
the detector shown in \Cref{fig:cwlr05_det}: it is
easier to detect adversarial samples from $2x^2+4x+5<6$ to attack $2x^2+3x+5<6$.

This indicates that, when the models get
regularized to polynomials that are similar, the attacks between
them end up being transferable.
Fortunately, the detection rates remain relatively high even in this case.
Different polynomials restrict the activation value range in different ways.
In practice, we found that the smaller the range that
is available , \ie~the unpenalized space for
activations, the harder it is to train the network.
Intuitively, if the regularization applies too strict a constraint to
activation values,
the stochastic gradient descent process struggles to converge
in a space full of local minima.
However, a strong restriction on activation values caused detection
rates to improve. Thus, when deploying
models with Sitatapatra, the choice of polynomials affects base
accuracies, detection rates and training time.

Key choice brings efficiency considerations to
the table as well. Although training time for individual devices
may be a bearable cost in many applications, a substantial increase in
run-time computational cost will often be unacceptable.
\Cref{tab:overhead} reflects
the total additional costs incurred
by using Sitatapatra in our evaluation models.
The \( < 7.5\% \) overhead we add
to the base network is small,
given that the models demonstrate
good defence and detection rates
against adversarial samples.
It is notable that during inference,
a detector module requires for each value
\( n \) fused multiply-add operations
to evaluate a \( n \)-degree \( f_k \) polynomial
using Horner's method~\cite{knuth1962evaluation},
and \( 1 \) additional operation for threshold comparison,
and thus utilize \( (n + 1) C_l H_l W_l \) operations
in total for the \( \ordinal{l} \) convolutional layer.
In our evaluation,
we set \( n = 2 \) by using second-order polynomials.
Additionally, a guard module
of the \( \ordinal{l} \) layer uses
channel-wise averaging, a fully connected layer,
a channel-wise scaling,
and element-wise multiplications for all activations,
which respectively require
\( C_l H_l W_l \), \( C_l^2 \), \( C_l \),
and \( C_l H_l W_l \) operations.

\subsection{Key attribution}
While exploring which activations ended up triggering an
alarm with different Sitatapatra parameters, we
noticed that we could often attribute
the adversarial samples to the models used to
generate them.
In practice, this is a huge benefit, known as {\em traitor tracing} in
the crypto research community: if somebody notices an attack,
she can identify the compromised device and take appropriate action.
We conducted a simple experiment to evaluate how to identify
the source of a different adversarial sample.
We first produced models instrumented with different
polynomials using Sitatapatra and then generated the
adversarial examples.

In this simple experiment, we use
a support vector machine (SVM) to classify
the adversarial images based on the models generated them. The
SVM gets trained on the 5000 adversarial samples and gets
tested on a set of 10000 unseen adversarial images.
The training and test sets are disjoint.

\begin{figure}[!ht]
\begin{minipage}{0.48\textwidth}
\centering
\begin{adjustbox}{scale=0.6}
\begin{tabular}{l|lllll}
    \toprule
    \multirow{2}{*}{Model}
        & \multicolumn{5}{c}{{\#FLOPs}} \\
    {}
        & original & detector & guard
        & total & overhead \\
    \midrule
    LeNet5
        & 480,500 & 28,160 & 7,660
        & 516,320 & 7.45\% \\
    ResNet18
        & 37,016,576 & 152,094 & 1,847,106
        & 39,015,776 & 5.40\% \\
    MCifarNet
        & 174,301,824 & 715,422 & 646,082
        & 175,663,328 & 0.64\% \\
    \bottomrule
\end{tabular}
\end{adjustbox}
\caption{%
    The computational costs,
    measured in the number of FLOPs,
    added by the detector and guard modules
    to the original network.
}\label{tab:overhead}
\end{minipage}
\hfill
\begin{minipage}{0.48\textwidth}
\centering
\begin{adjustbox}{scale=0.6}
\begin{tabular}{@{}l|lll|lll@{}}
\toprule
&\multicolumn{3}{c|}{FGSM}&\multicolumn{3}{c}{FGSM+DeepFool+C\&W}\\
Polynomial         & Precision & Recall & F1   & Precision & Recall & F1   \\\midrule
$0.1x^2-1x+2 < 3$  & \textbf{0.90}      & \textbf{0.90}   & 0.90 & 0.81      & 0.29   & 0.43 \\
$x^2-22x+120 < 10$ & \textbf{0.90}      & \textbf{0.89}   & 0.90 & \textbf{0.85}      & 0.28   & 0.42 \\
$2x^2+3x+5 < 6$    & 0.53      & 0.72   & 0.61 & 0.34      & 0.33   & 0.33 \\
$2x^2+4x+5 < 6$    & 0.56      & 0.35   & 0.43 & 0.30      & 0.69   & 0.41 \\ \midrule
Micro F1           & 0.72      & 0.72   & 0.72 & 0.40      & 0.40   & 0.40 \\
Macro F1           & \textbf{0.72}      & 0.72   & 0.71 & \textbf{0.57}      & 0.40   & 0.40 \\
\bottomrule
\end{tabular}
\end{adjustbox}
\caption{Key attribution based on the adversarial sample produced.}
\label{tab:key_attribution}
\end{minipage}
\end{figure}

\Cref{tab:key_attribution} shows the
classification results for FGSM-generated samples and all
attacks combined.
Precision, Recall and F1 scores are reported.
In addition, we report the micro and macro aggregate
values of F1 scores -- micro counts the total true positive,
false negatives and false positives globally, whereas macro
takes label imbalance into account.

For both scenarios, we get a performance that is much better than random
guessing. First, it is easier to attribute
adversarial samples generated by the large coarse-grained
attacks. Second, for polynomials that are different enough,
the classification precision is high -- for just FGSM we get a
$90\%$ recognition rate, while with more attacks it falls to $81\%$
and $85\%$. For similar polynomials, we get worse performance, of around
$50\%$ for FGSM and around $30\%$ for all combined.

This bring an additional trade-off -- training a lot of
different polynomials is hard, but it allows easier
identification of adversarial samples.
This then raises the question of scalability of the Sitatapatra model
instrumentation.

\subsection{Key size}

In 1883, the cryptographer Auguste Kerckhoffs enunciated a design principle that is used today: a system should withstand enemy capture, and in particular it should remain secure if everything about it, except the value of a key, becomes public knowledge~\cite{kerchoffs1883}.
Sitatapatra follows this principle -- as long as the key is secured, it will, at a minimal cost, provide an efficient way to protect against
low-cost attacks at scale. The only difference is that if an opponent
secures access to a system protected with Sitatapatra, then by observing
its inputs and outputs he can use standard optimisation methods to
construct adversarial samples. However these samples will no longer
transfer to a system with a different key.
One of the reasons why sample transferability is an
unsolved problem is because it has so far been hard to generate a wide
enough variety of models with enough difference between
them at scale.

Sitatapatra is the first attempt at this problem, and it shows
possibilities to embed different polynomial functions as keys to CNNs to block
transfer of adversarial samples.
Modern networks have hundreds of layers and channels, and thus can embed different keys at various parts improving key diversity.
The potential implication of having a key is important.
In the case where a camera vendor wants to sell the same vision system to
banks as in the mass market, it may be sufficient to have a completely
independent key for the cameras sold to each bank.
It will also be sensible for a mass-market vendor wants to
train its devices in perhaps several dozen different families, in order
to stop attacks scaling (a technique adopted by some makers of pay-TV
smartcards~\cite{And2008}). However it would not be practical to have an
individual key for each unit sold for a few dollars in a mass market.
The fact that adding keys to neural networks can improve its diversity
solves the problem of training multiple individual neural networks on the
same dataset, which is also known as a model ensemble.
In fact, our key instrumentation and the model identification from a model ensemble together can be seen as a stream of deployment, which also sufficiently increases the key space.

\section{Conclusion}\label{sec:conc}

In this paper we presented Sitatapatra, a new way to use both static and
dynamic channel manipulations to stop adversarial sample transfer.
We show how to equip models with guards that diffuse gradients and
detectors that restrict their ranges, and demonstrate
the performance of this combination on CNNs of varying sizes.
The detectors enable us to introduce key material as in cryptography
so that adversarial samples generated on a network with one key will
not transfer to a network with a different one, as activations will
exceed the ranges ranges permitted by the detectors and set off an
alarm. 
We described the trade-offs in terms of accuracy, detection
rate, and the computational overhead both for training and at run-time.
The latter is about five percent, enabling Sitatapatra to work on
constrained devices. The real additional cost is in training but in
many applications this is perfectly acceptable. Finally, with a proper
choice of transfer functions, Sitatapatra also allows adversarial sample
attribution, or `traitor tracing' as cryptographers call it -- so that a
compromised device can be identified and dealt with.

\section*{Acknowledgements}
\textit{Partially supported with funds from Bosch-Forschungsstiftung
im Stifterverband.}

\bibliography{references}
\bibliographystyle{abbrv}

\end{document}